# An Educational Fuzzy-based Control platform using LEGO Robots

Daniel Zaldivar, Erik Cuevas, Marco A. Pérez-Cisneros, Juan H. Sossa, Jose G. Rodríguez, Edgar O. Palafox

[a]Departamento de Electrónica, Universidad de Guadalajara, CUCEI
Av. Revolución 1500, Guadalajara, Jal, México

**Abstract**

Fuzzy controllers have gained popularity in the past few decades with successful implementations in many fields that have enabled designers to control complex systems through linguistic-based rules in contrast to traditional methods. This paper presents an educational platform based on LEGO© NXT to assist the learning of fuzzy logic control principles at undergraduate level by providing a simple and easy-to-follow teaching setup. The proposed fuzzy control study aims to accompany students to the learning of fuzzy control fundamentals by building hands-on robotic experiments. The proposed educational platform has been successfully applied to several undergraduate courses within the Electronics Department in the University of Guadalajara. The description of robotic experiments and the evaluation of their impact in the student performance are both provided in the paper.

## 1. Introduction

Traditional methods to control any dynamic system require the use of some knowledge about the model of the system to be controlled. An accurate approximation of the system is thus crucial for the successful implementation of the control algorithm. Unfortunately, most real systems are nonlinear, highly complex, and too difficult or impossible to model accurately. Fuzzy logic is a mathematical framework developed by Lofti Zadeh [1] which helps to reduce the complexity of controlling nonlinear systems. Fuzzy logic expresses operational laws of a control system in linguistic terms instead of traditionally used mathematical equations. Such linguistic terms are expressed in the form of IF-THEN rules. Among several successful fuzzy control approaches in the literature, the fuzzy PD controller [2-6] is one of the most robust and easy to implement algorithms.

Fuzzy control is an elective course at the second year of Electronic Engineering. The course attracts about 20 students per semester at each academic year. It has been agreed among our faculty members that the introduction of fuzzy control concepts to undergraduates is challenging and that experimental components such as an educational platform radically helps to understand the practical implications of fuzzy control. However, the number of commercially available educational fuzzy control laboratory kits is rather limited and the existing ones are either too expensive or relative complex to be used. A quick overview on some current available fuzzy logic laboratory kits that offer practical hands-on experience to students is discussed below.

GUNT [7] provides an introduction to digital real-time control by fuzzy methods. The setup includes a ball-beam model which acts as a mechanical single-variable system. A fuzzy control is used to attempt to hold the ball in a specific position by tilting the beam. Control algorithms are initially written in the development software FSH-Shell and then compiled to generate microcontroller code. However, its relative high cost and large dimensions have yielded a non-practical individual development setup which should be rather considered as a demonstrative tool.

On the other side, the CE124 [8] has been designed to allow users to quickly implement fuzzy logic in an intuitive way. It provides a schematic panel compound with two fuzzifiers, a fuzzy state table, a set of fuzzy logic gates AND & OR, NOT blocks and a defuzzifier. It is a unique hardware module whose orientation aims to implement fuzzy systems by using analogue elements. Despite delivering a quick implementation, it does not provide convenient programming properties which are normally available for digital based systems.

FuzzyTECH [9] is a family of fuzzy logic and neural-fuzzy development tools whose programming is completely graphical and based on ISO 9000 and IEC 1131-7 fuzzy logic standards. It provides code and

[Escriba texto]

runtime modules for microcontrollers, PLCs, process control systems and popular software packages such as InTouch, LabVIEW, FIX, and Matlab/Simulink. A special version, fuzzyTECH for Business is targeted at financial and commercial applications of fuzzy logic and neuro-fuzzy technologies. However, most of them are relatively complex to test and more oriented as a final application tool.

The Fuzzy Logic Toolbox for MATLAB [10] is a set of easy-to-use functions for building and evaluating fuzzy logic systems. Thousands of code examples are available since world-wide users have made contributions. However, the Matlab© license is very expensive and requires users to be familiar with the MATLAB programming environment.

It is important to consider that educational goals can be achieved by a wide range of complexity, hardware, software, hardware-software, sizes and costs in a variety of platforms. Fuzzy control theory and applications may be learnt through paper-and-pencil exercises, traditional computer programming or complemented by robotics exercises. This work has been developed under the hypothesis that the use of robotics motivates the learning of fuzzy control by connecting theory and exercises yielding a deeper and lasting impact on the student.

So far, the use of robotics has been positive to encourage the student's interest on engineering topics [11-19]. They can quickly become familiar to the platform and test their own ideas based on project designing with conducted experiments that strength theoretical concepts. This expands student knowledge beyond their immediate interests demonstrating the benefits of hands-on robotics on the classroom [20]. It is therefore not surprising that many universities are incorporating the use of robotics plants to teach several engineering subjects.

The LEGO© NXT is a friendly, affordable, motivational and effective educational robotics platform [21] which has demonstrated its use for research and educational proposes [22-29]. At first glance, the LEGO© components offer interesting possibilities. Mindstorms LEGO kits include relatively low-cost hardware and software which are suitable for implementing complex designs for Robotics, Mechatronics and Control. Students can become quickly familiar with the set by taking further steps and adventuring to the designing and testing their own add-on components. They might also interface the kit to other processing and sensing units such as microcontrollers, sensors, etc.

In this paper, a simple educational platform based on LEGO© NXT is presented to assist the learning of fuzzy control concepts at undergraduate level. The course starts by theoretically introducing the fuzzy PD algorithm to the students. The capabilities and limitations of LEGO© NXT mobile robots are studied with two experiments using the fuzzy PD controller which are timely provided by the lecturer. The first project concerns about the robot's orientation while the second solves a robotic path tracking problem. Students are encouraged to code the real-time application by using their own programming skills over the LEGO platform using ROBOTC©. During the whole process, students strongly reinforce the learning of the fuzzy control issues as their motivation is encouraged by the robotic application and the visual results that emerged from experiments and the programming experience. The paper also discusses on the positive impact of robotics hands-on experiments for motivating students' interest on learning the fuzzy control theory.

The paper is organized as follows. Section 2 explains the fuzzy PD controller algorithm which features in the experiments. Section 3 shows the LEGO System configuration. Section 4 shows the robot's orientation project while Section 5 presents the path tracking setup. Section 6 discusses on the student's performance and its evaluation as some conclusions and contributions of the paper are drawn in Section 7.

## 2. The fuzzy PD controller

In this work, the fuzzy PD controller [2-6] is used to implement robot experiments. It consists of only 4 rules and has the structure illustrated in Figure 1.

[Escriba texto]

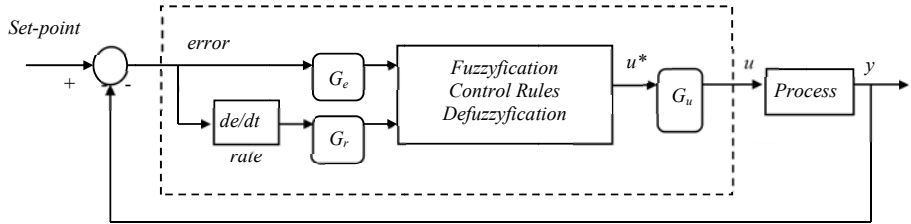

**Figure 1** Fuzzy PD controller structure

The values *Gu, Ge* and *Gr* correspond to the output gain, the error gain and error rate gain respectively. All are to be determined either by experimentation or by the method described on [30]. The value *u\** is the defuzzyficated output, which is also known as "crisp output".

### 2.1 Fuzzyfication

As it is shown in Figure 1, there are two inputs to the controller: *error* and *rate* (the error's change). The *error* is defined as:

$$error = setpoint - y \qquad (1)$$

The *rate* is defined as follows:

$$rate = (ce - pe) / sp, \qquad (2)$$

where *ce* is the current error, *pe* is the previous error and *sp* is the sampling period. Figure 2 represents two membership functions (positive and negative) for each input (error and rate), with two pairs of membership functions, one at each input (one for the error and one for the rate) and four input membership functions: error positive (ep), error negative (en), rate positive (rp) and rate negative (rn). The output membership function is shown in Figure 3. H and *L* are two positive constants to be determined by experimentation [4-6]. In order to reduce the number of control parameters, it is assumed that H=L. It is observed that its value depends on the limit of universe of discourse for the input variable of each specific control problem as it is presented in Section 4.1.

The membership functions for the input variables, *error* and *rate*, are defined as described in [2-3] as follows:

$$\mu_{ep} = \frac{L+(G_e * error)}{2L} \qquad \mu_{en} = \frac{L-(G_e * error)}{2L}$$

$$\mu_{rp} = \frac{L+(G_r * rate)}{2L} \qquad \mu_{rn} = \frac{L-(G_r * rate)}{2L} \qquad (3)$$

Were $\mu_{ep}$ is the error positive, $\mu_{en}$ is the error negative, $\mu_{rn}$ is the rate negative, and $\mu_{rp}$ is the rate positive.

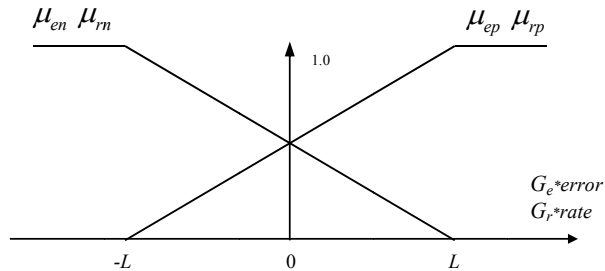

**Figure 2** Input membership functions

[Escriba texto]

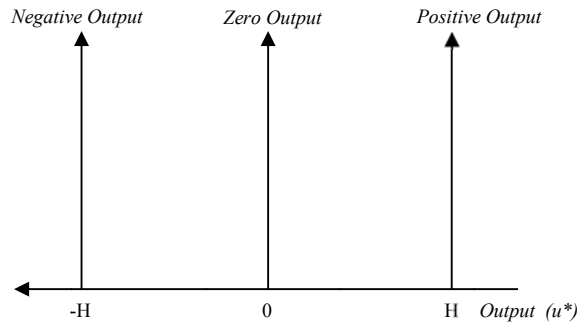

**Figure 3** Output membership functions

### 2.2 Fuzzy rules and defuzzyfication

According to [4] there are four rules to evaluate the fuzzy PD controller:

R1. If *error* is **ep** and *rate* is **rp** then output is **op**

R2. If *error* is **ep** and *rate* is **rn** then output is **oz**     (4)

R3. If *error* is **en** and *rate* is **rp** then output is **oz**

R4. If *error* is **en** and *rate* is **rn** then output is **on**

Above, **ep** is the positive error, **en** is the negative error, **rn** is the negative rate and **rp** is the positive rate. More details about the interpretation of these four rules are provided by [4].

According to [3], the defuzzyfication method employs the centre of mass method as follows:

$$u = \frac{-H(\mu_{R4(x)}) + 0(\mu_{R2(x)+}\mu_{R3(x)}) + H(\mu_{R1(x)})}{\mu_{R4(x)} + (\mu_{R2(x)+}\mu_{R3(x)}) + \mu_{R1(x)}} \quad (5)$$

Where $\mu_{R1(x)}, \mu_{R2(x)}, \mu_{R3(x)}, \mu_{R4(x)}$ are the membership input values and $-H, 0, H$ are the membership output values.

In order to simplify the controller algorithm and save computing time, the fuzzy PD controller, adopted at this work, follows the design proposed in [4-5]. In such approach, the input space is divided into a 20-inputs region (IC) as it is shown by Figure 4. Nine equations are used to calculate the control output *u* (see Equation 6) with only one being used depending on which region is activated by the current combination of the *error* value and its *rate*.

[Escriba texto]

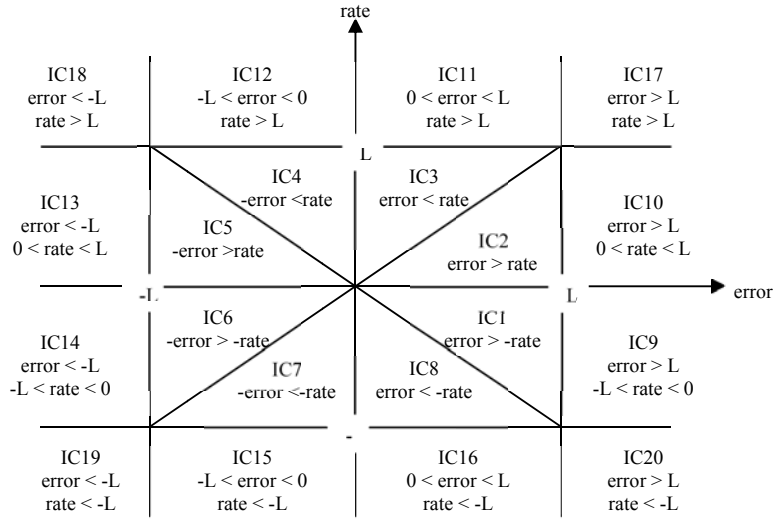

**Figure 4** Input combinations

$$u = \frac{L}{2(2L - G_e |error|)}[G_e * error + G_r * rate] \quad \text{for } IC1, IC2, IC5, IC6.$$

$$u = \frac{L}{2(2L - G_r |rate|)}[G_e * error + G_r * rate] \quad \text{for } IC3, IC4, IC7, IC8.$$

$$u = \frac{1}{2}[L + G_r * rate] \quad \text{for } IC9, IC10. \tag{6}$$

$$u = \frac{1}{2}[L + G_e * error] \quad \text{for } IC11, IC12.$$

$$u = \frac{1}{2}[-L + G_r * rate] \quad \text{for } IC13, IC14.$$

$$u = \frac{1}{2}[-L + G_e * error] \quad \text{for } IC15, IC16.$$

$$u = L \quad \text{for } IC17.$$

$$u = -L \quad \text{for } IC19.$$

$$u = 0 \quad \text{en } IC18, IC20.$$

## 3. Experimental setup

The block diagram for the LEGO system for both exercises is shown in Figure 5. The system uses two servo motors connected to out ports A and B as actuators. Such motors are controlled by a PWM signal (already embedded and transparent to the user) which is generated by the LEGO brick. On the other side, a compass sensor from HiTechnic© is used as input signal in port 1. Such sensor is a digital compass that measures the earth's magnetic field and outputs a value representing the current heading. The magnetic heading is calculated to the nearest 1° and returned as a number from 0 to 359. This third-part sensor is totally prepared to be compatible with the LEGO NXT© electronics, whose implementation using the ROBOTC© program is transparent for the user.

[Escriba texto]

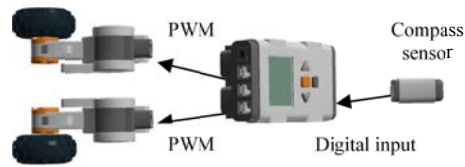

**Figure 5.** Diagram of the LEGO system

## 4. Orientation control using the Fuzzy PD controller

In this section, the experiment to control the robot's orientation $\theta_{ref}$ using the fuzzy PD controller is explained. In order to reach the reference angle $\theta_{ref}$, the angle's error $e_\theta$ must be calculated by $\theta_{ref} - \theta_{actual}$ as it is depicted by Figure 6. Considering the fuzzy approach presented on Section 2, the orientation control problem is handled as follows: Using as inputs the angle's error $e_\theta$ (see Figure 7) and the angle's error rate $r$ ($de_\theta / dt$), the fuzzy PD controller identifies the region (input combination) which corresponds to one out of 20 possible regions (see Figure 4). Once the control region has been defined, one of the nine Eq. (6) is used to calculate the controller output $u$.

### 4.1 Implementation

This section starts by explaining the implementation of the first experiment. Algorithm 1 depicts the whole strategy for controlling the robot's orientation. Line 1 shows the controller's gains which have been experimentally obtained and the L value. Since the angle varies from 360 to -360 degrees, the positive limit (L) is set to 360, defining the limits of interest for the universe of discourse at this problem. Using ROBOTC© as development platform, students are encouraged to code the real-time control software.

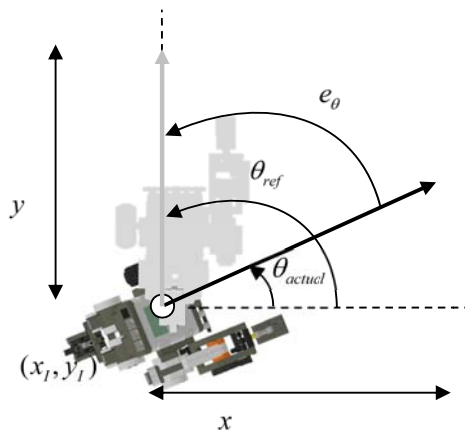

**Figure 6.** Obtaining the angle's error $e_\theta$

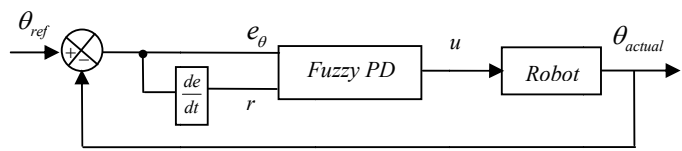

**Figure 7.** Robot's orientation control

[Escriba texto]

Orientation control Algorithm

```
    G_e : error's gain
    G_r : rate's gain
    G_u : output's gain
    L: universe of discourse limit
    θ_ref : desired angle in degrees
    θ_actual : actual angle in degrees
    u: output of the fuzzy controller
    r: rate of the angle's error
    e_θ : angle's error
    e_{θ-1} : angle's error obtained from previous period of time

1:  G_e ← 20,  G_r ← 2, G_u ← 0.5    • defining the controller's gains
2:  L ← 360                          • and its positive limit
While (true)
3:  e_θ ← θ_ref − θ_actual           • obtaining the angle's error
4:      if( e_θ > π ) then           • checking if the error is bigger than 180
degrees, to turn the robot clockwise
5:  e_θ ← e_θ -2 π
6:      if( e_θ < - π ) then         • checking if the error is smaller than -180
degrees, to turn the robot anti-clockwise
7:  e_θ ← e_θ +2 π
8:  r ← e_θ − e_{θ-1}                • obtaining the angle's error rate
9:  e_{θ-1} ← e_θ                    • obtaining the last angle's error
10: e_θ ← e_θ * G_e                  • multiplying the actual angle's error by its gain
11: r ← r * G_r                      • multiplying the rate by its gain
12: u ← FUZZY( e_θ ,r)               • calling the fuzzy function to obtain the output
13:     u ← u * G_u                  • multiplying the output by its gain
14:     if ( e_θ >0) then             • If the angle's error is positive then the
robot rotates anticlockwise
15:         motorA ← u
16:         motorB ← -u
17:         else ( e_θ <0)            • If the angle's error is negative then the
robot rotates clockwise
18:         motorA ← -u
19:         motorB ← u
```

**Algorithm 1.** Orientation control algorithm

In order to eliminate the error $e_\theta$ while efficiently reaching $\theta_{ref}$, lines 4 and 6 of the Algorithm 1 have been included to define the shortest way to meet the reference angle considering the actual robot position. Figure 8 shows the inclusion of such code through a generic example.

Considering its importance, the fuzzy PD controller is described as a separate function (FUZZY($e_\theta$, $r$)) which is shown in the Algorithm 2.

Fuzzy PD Algorithm

```
1: FUZZY( e_θ ,r)

2: Evaluate the 20 regions to find the output u

3: if ( e_θ >-r) or ( e_θ >r) or (- e_θ >r) or (- e_θ >-r) then    • regions ic1, ic2, ic5, ic6
4:      u ← (L/(2*(2*L-abs( e_θ ))))*( e_θ +r)

5: if ( e_θ <r) or (- e_θ <r) or (- e_θ <-r) or ( e_θ <-r) then    • regions ic3, ic4, ic7, ic8
6:      u ← (L/(2*(2*L-abs(r))))*( e_θ +r)

7: if (( e_θ >L) and (-L<r<0)) or ((0<r<L) and( e_θ >L)) then      • regions ic9, ic10
8:      u ← (L+r)/2

9: if ((0< e_θ <L) and (r>L)) or ((r>L) and(-L< e_θ <0)) then      • regions ic11, ic12
10:     u ← (L+ e_θ )/2

11: if ((e<-L) and(0<r<L)) or ((e<-L) and (-L<r<0)) then           • regions ic13, ic14
12:     u ← (-L+r)/2
```

[Escriba texto]

```
13: if ((r<-L) and (-L<e_θ<0)) or ((r<-L) and (0<e_θ<L)) then    • regions ic15, ic16
14:       u • (-L+e_θ)/2
15: if (e_θ>L) and (r>L) then                                    • regions ic17
16:       u ← L
17: if (e_θ<-L) and (r<-L) then                                  • regions ic19
18:       u ← -L
19: if ((e_θ<-L) and (r>L)) or ((e_θ>L) and (r<-L)) then          • regions ic18, ic20
20:       u • 0
21:    return u
```

**Algorithm 2.** Fuzzy PD controller function

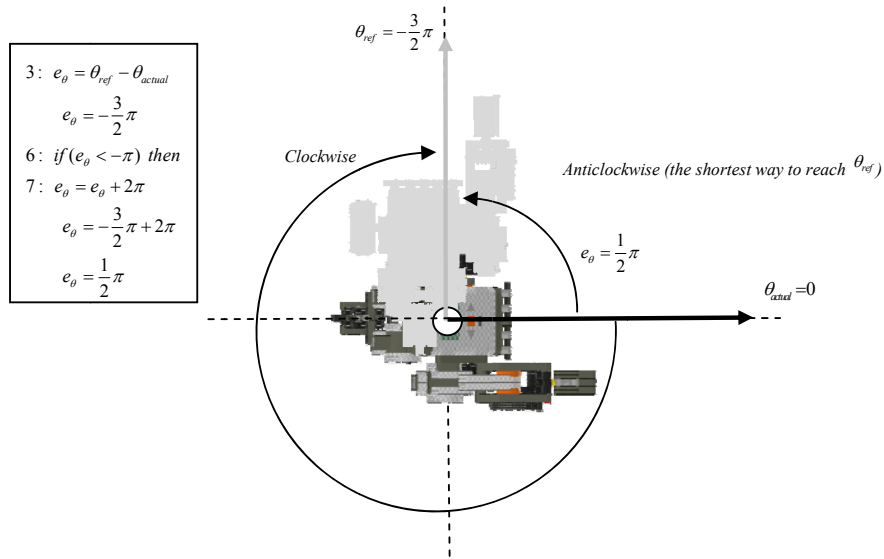

**Figure 8.** Reaching efficiently $\theta_{ref}$, (i.e. deciding if it does spin clockwise or anticlockwise).

Figure 9 shows the performance of the real-time fuzzy controller for orientation, considering an initial $e_\theta$ of 90 degrees. Figure 9, below, shows the evolution of power for the left ($\omega_l$) and right ($\omega_r$) motors over 300 iterations. Such graphic is a compulsory assignment for each student during the course.

[Escriba texto]

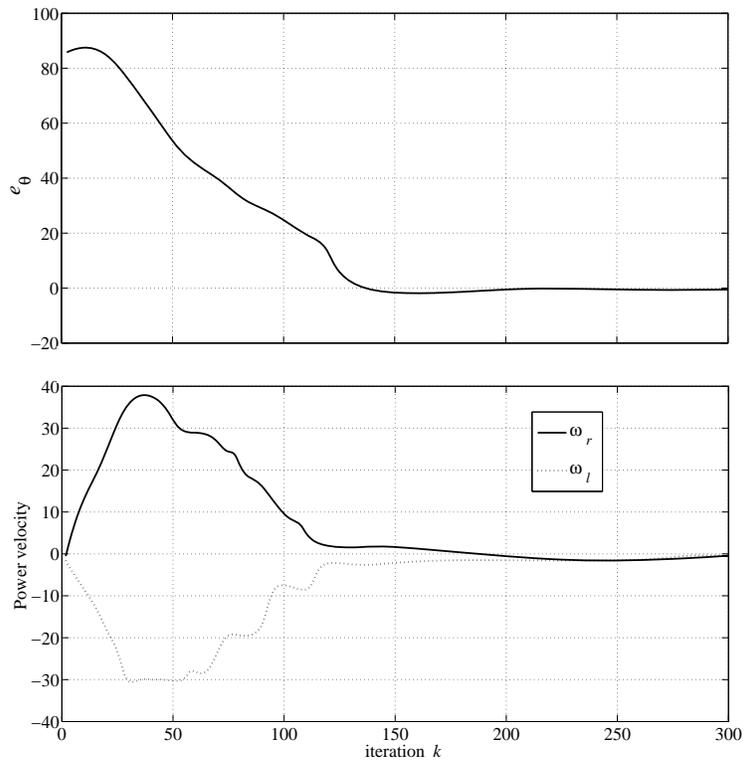

**Figure 9.** Performance of the real-time fuzzy controller for the orientation problem

## 5. The path tracking experiment

This section presents an experiment to control the tracking of a given robot's path using a fuzzy PD controller. Figure 10 shows the path tracking control procedure step by step as follows: 1.- The robot's actual position ($x_{actual}$, $y_{actual}$) and the goal position ($x_{ref}$, $y_{ref}$) are used in order to obtain the goal angle $\theta_{ref}$, and the relative robot's distance $R$ to the goal position. 2. $R$ is used to regulate the linear velocity $v$ by considering a linear gain $K_p$. 3. $\theta_{ref}$ and $\theta_{actual}$ (obtained by a compass sensor) are used to obtain $e_\theta$ and $r$, the angle error's and its derivate, respectively. 4. Both signals ($e_\theta$ and $r$) are used as inputs to the fuzzy PD controller in order to obtain the robot's angular velocity $\omega$. 5. Since the robot is controlled only by the angular velocity at each wheel, the linear velocity $v$ and the angular velocity $\omega$ are used to obtain velocities $\omega_l$ and $\omega_r$ according to their relationship which is explained in the robot's model [29]. Lines 28 and 30 of the algorithm 3 show the procedure.

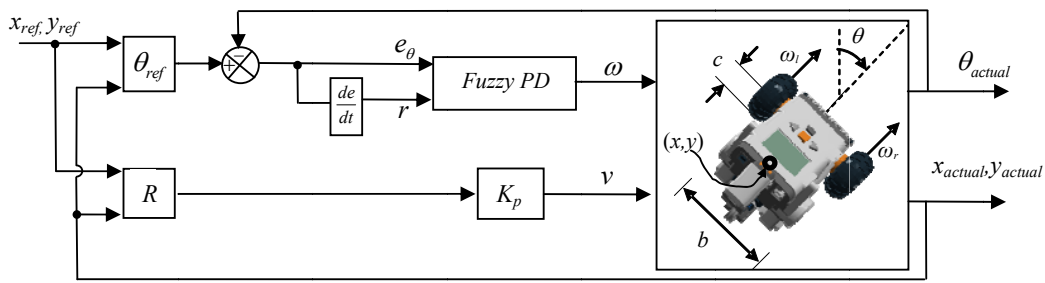

**Figure 10.** Robot's path tracking control

[Escriba texto]

### 5.1 Implementation

In the Algorithm 3, the general procedure to implement the path tracking controller is exposed as pseudo code. Line 1 shows the controller's gains which have been experimentally obtained and the L value. This algorithm is explained to the students and they are encouraged to code it by themselves in ROBOTC©.

**Path-tracking Control Algorithm**

$G_e$, $G_r$, $G_u$ : error, rate and output gains

$\theta_{ref}, \theta_{actual}$ : desired and actual angles in degrees

$u$: output of the fuzzy controller

$e_\theta, e_{\theta-1}$ : angle's error and the last angle's error

$r$: rate of the angle's error

$x_{actual}, y_{actual}$ : actual coordinates

$x_{ref}, y_{ref}$ : reference coordinates

$R$ : robot's distance to the next point

$v, w$ : lineal and angular velocities

$K_p$ : proportional controller's gain

$w_l, w_r$ : left and right angular velocities

1: $G_e \bullet 4.7$, $G_r \bullet 0.025$, $G_u \bullet 1.01$, $L \bullet 360$ • defining gains for the fuzzy controller
   $K_p \bullet 0.25$                                                 • defining the proportional controller's gain

2: $b \bullet 0.094$, $c \bullet 0.028$, $PER \bullet 0.179$ • defining robot's constants (b length of the back shaft, c wheel's radius, and PER accounts for the robot´s wheel perimeter)

3: $x_{ref} \bullet 1.0$, $y_{ref} \bullet 1.0$          • defining the robot's reference coordinates

**While** (true)
4:  $tc_r \bullet motor\_get\_count(portB)$        • reading the right wheel robot's tick counter
5:  $tc_l \bullet motor\_get\_count(portA)$        • reading the left wheel robot's tick counter

6:  $rot_r \bullet tc_r * PER / 360$               • calculating the right wheel robot's rotations
7:  $rot_l \bullet tc_l * PER / 360$ :             • calculating the left wheel robot's rotations
8:  $rot_{ave} \bullet rot_r + rot_l / 2$          • both wheels robot's rotations are averaged
9:  $\delta_\theta \bullet rot_r - rot_l / b$      • obtaining the angle's increment
10: $\delta_y \bullet rot_{ave} \times \sin(\theta_{actual} + \delta_\theta / 2)$   • obtaining the distant increment on y
11: $\delta_x \bullet rot_{ave} \times \cos(\theta_{actual} + \delta_\theta / 2)$   • obtaining the distant increment on x

12: $x_{actual} \bullet x_{actual} + \delta_x$     • calculating the $x_{actual}$
13: $y_{actual} \bullet y_{actual} + \delta_y$     • calculating the $y_{actual}$
14: $\theta_{actual} \bullet \theta_{actual} + \delta_\theta$  • calculating the $\theta_{actual}$

15: $x_e \leftarrow x_{ref} - x_{actual}$          ▶ calculating the error on x
16: $y_e \leftarrow y_{ref} - y_{actual}$          ▶ calculating the error on y

17: $\theta_{ref} \leftarrow \tan^{-1} \frac{y_e}{x_e}$   ▶ calculating $\theta_{ref}$

18: $R \leftarrow \sqrt{x_e^2 + y_e^2}$            ▶ calculating R

19: $e_\theta \leftarrow \theta_{ref} - \theta_{actual}$   ▶ obtaining the angle's error
20: $r \leftarrow e_\theta - e_{\theta-1}$         ▶ obtaining the angle's error rate
21: $e_{\theta-1} \leftarrow e_\theta$             ▶ obtaining the last angle's error
22: $e_\theta \leftarrow e_\theta * G_e$           ▶ multiplying the actual angle's error by its gain
23: $r \leftarrow r * G_r$                         ▶ multiplying the rate by its gain

24: $u \leftarrow \text{FUZZY}(e_\theta, r)$       ▶ calling the fuzzy function
25: $u \leftarrow u * G_u$                         ▶ multiplying the controller's output by its gain

[Escriba texto]

```
26:   if(u> L)then    ▶ preventing an overflow of the output by limiting its value to
fall between L and -L
              u= L

      if(u< L)then
              u= -L

27:   v ← R×K_p    ▶ robot's lineal velocity scaled by a linear gain.

28:   w_l = (v - u×(b/2))×(1/c)    ▶ obtaining the left wheel velocity

29:   w_l = 1 - exp^{-0.8 w_l}     ▶ avoiding saturation

30:   w_r = (v + u×(b/2))×(1/c)    ▶ obtaining the right wheel velocity

31:   w_r = 1 - exp^{-0.8 w_r}     ▶ avoiding saturation

32:          motorA ← w_l    ▶ applying the output to the robot's left motor

33:          motorB ← w_r    ▶ applying the output to the robot's right motor
```

**Algorithm 3.** Path tracking control algorithm

Figure 11 shows the performance of the fuzzy controller for a generic path-tracking task. The robot's reference trajectory and the actual trajectory are both plotted in centimeters.

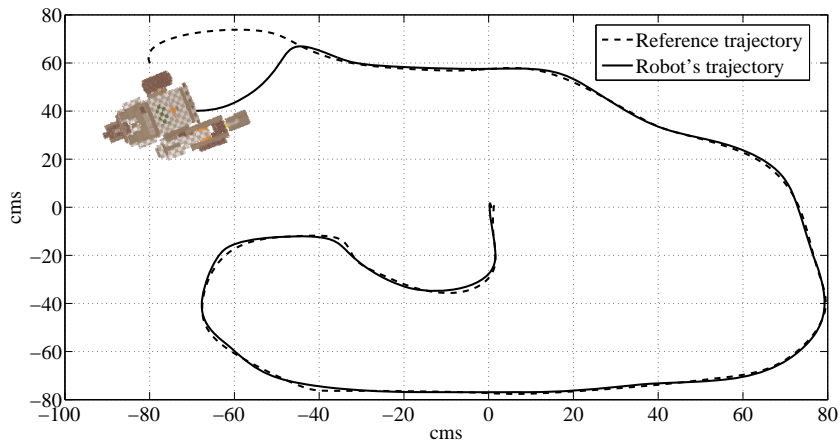

**Figure 11.** The fuzzy controller performance over a path-tracking task.

## 6. Results and discussion

This section describes the positive impact on students after the implementation of hands-on robotic exercises whose target has been the motivation for learning fuzzy control theory. All the experiments in this paper have been incorporated into an elective second-year undergraduate course taught in the Electronics Department of the University of Guadalajara.

This work considers two performance indexes: the *Course Examination* and the *Examination after a year*. The former refers to an examination that has been applied to student just after finishing courses on fuzzy logic concepts. On the other hand, the latter considers a second examination that is applied to students after one year of finishing the fuzzy course. Such procedure aims to evaluate the student's knowledge retention after a year for both the practical and non-practical course attendees.

Both indexes consider a universe of forty students with twenty corresponding to those who attended a traditional fuzzy control course and the remaining referring to those students who have attended the fuzzy control hands-on robotic subject. Table 1 shows the values of the *Course Examination* and the *Examination after a year* in terms of the *Average* and the *Worst-Best* obtained values. The *Best* value, correspond to the highest mark while the *Worst* value corresponds to the lowest score.

[Escriba texto]

Table 1 presents the *Course Examination* index which shows that the marks for students attending the hands-on robotic course are higher than those given to students attending a traditional course. A similar trend is observed for the *Examination after a year* index. Again, the *Best* index shows that higher marks have been given to students attending the fuzzy hands-on control course. On the other hand, the *Worst* index addresses lower values for students attending a traditional fuzzy control course. Same trend has been observed for either the *Course Examination* or the *Examination after a year* index.

|  | Fuzzy control course | | | Fuzzy control course + hands-on robotic exercises | | |
|---|---|---|---|---|---|---|
|  | *Average* | *Best* | *Worst* | *Average* | *Best* | *Worst* |
| *Course Examination* | 70 | 92 | 60 | 82 | 100 | 73 |
| *Examination after a year* | 55 | 76 | 45 | 72 | 97 | 62 |

**Table 1.** The averaged scholar-performance results

### 7. Conclusions

Fuzzy controllers have become popular in recent decades with successful implementations in several industrial and academic subjects. Such trend has indeed triggered a remarkable availability of fuzzy control courses in many technical colleges and universities around the world.

By using fuzzy logic control, designing a controller is relatively simple task yet facing complex and nonlinear systems. In this paper, the authors have presented the use of a simple educational platform based on LEGO© NXT to assist the learning of fuzzy control concepts in undergraduate engineering courses. The capabilities and limitations of LEGO© NXT mobile robots have been studied within two experiments using a fuzzy PD controller. The first experiment addresses the robot's orientation problem while the second solves the robot's path tracking task. Both algorithms and their required additional tools are presented including their corresponding experimental results. The paper has also shown the potential of a LEGO© NXT as a robust robotic platform for teaching fuzzy control. The platform enables students to implement control algorithms by simply doing programming at a high-level language interface. Experimental evidence shows an increase in the performance and reinforcement of learning fuzzy control issues for the students who have attended the hands-on robotic exercises. Similarly, evidence shows a trend to hold the knowledge of fuzzy control issues for longer periods of time for those students who have worked through the proposed exercises. Such results indicates that the use of robotics motivates the learning of fuzzy control by connecting theory and exercises yet sowing a deeper and lasting impact on the students.

**References**

[1] L. A. Zadeh, ´Fuzzy sets´. *Information and control*, Vol. 8, Academic, New York, 1965, pp 338-353.

[2] H. Ying, W. Siler, J. J. Buckley, ´Fuzzy Control Theory: A Nonlinear Case, *Automatica*, **26** (3) 1990, 513-520.

[3] H.A. Malki, h. Li, G. Chen, ´New design and stability analysis of fuzzy proportional-derivative control systems´, IEEE Transactions on Fuzzy Systems, **2** (1994), 245 – 254.

[4] Y.C Hsu, G. Chen, E. Sanchez, 'A Fuzzy controller for a Multi-Link Robot Control: Stability Analysis´ *Proceedings of the 1997 IEEE International Conference on Robotics and Automation*, Albuquerque, NM, 1997, pp. 1412-1417.

[5] E. Sanchez, L.A. Nuno, Y.C. Hsu, G. Chen, ´Fuzzy PD scheme for underactuated robot swing-up control´, *Fuzzy Systems Proceedings, 1998. IEEE World Congress on Computational Intelligence., The 1998 IEEE International Conference on* , **1**, (1998), 302-306.

[6] F. Lara-Rojo, E.N. Sanchez, D. Zaldivar, ´Minimal fuzzy microcontroller implementation for didactic applications´, *Journal of Applied Research and Technology*, **1**(2) (2003) 137-147.

[Escriba texto]

[7] GUNT Geratebau, GmbH, RT 121 Fuzzy Control: Ball-on-Beam User's Guide, G.U.N.T. Geratebau, GmbH, Fahrenberg, Barsbuttel, 2006.

[8] CE124 Fuzzy Logic Trainer, http://www.control-systems-principles.co.uk/whitepapers/fuzzy-logic-systems.pdf [Visited on April 2012].

[9] FuzzyTech, http://www.fuzzytech.com/ [Visited on April 2012].

[10] Fuzzy Logic Toolbox for MATLAB, http://www.mathworks.com/products/fuzzy-logic/index.html;jsessionid=6524973bb2b9c817e0e0861001f8 [Visited on April 2012].

[11] L. Meeden. Using Robots As Introduction to Computer Science. In John H. Stewman, editor, *Proceedings of the Ninth Florida Artificial Intelligence Research Symposium (FLAIRS)*, pages 473--477. Florida AI Research Society, 1996.

[12] C. A. Jara, F. A. Candelas, S. T. Puente, F. Torres, ´Hands-on experiences of undergraduate students in Automatics and Robotics using a virtual and remote laboratory´, *Computers & Education*, **57** (2011), 2451–2461.

[13] J. M. Gómez-de-Gabriel, A. Mandow, J. Fernández-Lozano, and A. J. García-Cerezo, IEEE, ´Using LEGO NXT Mobile Robots With LabVIEW for Undergraduate Courses on Mechatronics´, IEEE TRANSACTIONS ON EDUCATION, **54**, (2011), 41-47.

[14] T. Karp, R. Gale, L. A. Lowe, V. Medina, and E. Beutlich ´Generation NXT: Building Young Engineers With LEGOs´, IEEE TRANSACTIONS ON EDUCATION, **53**, (2010), 80-87.

[15] A. Behrens, L. Atorf, R. Schwann, B. Neumann, R. Schnitzler, J. Ballé, T. Herold, A. Telle, T. G. Noll, K. Hameyer, and T. Aach, ´MATLAB Meets LEGO Mindstorms—A Freshman Introduction Course Into Practical Engineering´, IEEE TRANSACTIONS ON EDUCATION, **53**, (2010), 306-317.

[16] Beer, R. D.; Chiel, H. J.; and Drushel, R. F. 1999. Using autonomous robotics to teach science and engineering. *Communications of the ACM* 42(6).

[17] Kumar, D., and Meeden, L. 1998. A robot laboratory for teaching artificial intelligence. In *Proceedings of the 29th SIGCSE Technical Symposium on Computer Science Education (SIGCSE-98)*, volume 30(1) of *SIGCSE Bulletin*, 341–344. New York: ACM Press.

[18] Fabiane Barreto Vavassori Benitti, ´Exploring the educational potential of robotics in schools: A systematic review´, *Computers & Education*, **58** (2012), 978–988.

[19] P. Ranganathan, R. Schultz, M. Mardani, ´Use Of Lego Nxt Mindstorms Brick In Engineering Education´, *Proceedings of the 2008 ASEE North Midwest Sectional Conference*, 2008.

[20] L. Greenwald, J. Kopena, ´Mobile Robot Labs: on achieving educational and research goals with small, low-cost plataforms´, *IEEE Robotics and Automation Magazine*, **10**(2) (2003) 25-32.

[21] R. Ramli, M. M. Yunus and N. M. Ishak, ´Robotic teaching for Malaysian gifted enrichment program´, *Procedia Social and Behavioral Sciences*, **15** (2011), 2528–2532.

[22] D. Martinec and Z. Hurak, ´Vehicular Platooning Experiments with LEGO MINDSTORMS NXT´, in the IEEE International Conference on Control Applications (CCA), Part of 2011 IEEE Multi-Conference on Systems and Control, Denver, CO, USA. September, 2011, pp. 28-30.

[23] D. Zermas, Control of a Leader–Follower Mobile Robotic Swarm Based on the NXT Educational LEGO Platform, IEEE International Symposium on Industrial Electronics (ISIE), 2011, Rio, Greece, 2011, pp. 1381-1386.

[24] A. Vourvopoulos, F. Liarokapis, Brain-controlled NXT Robot: Tele-operating a robot through brain electrical activity, 2011 Third International Conference on Games and Virtual Worlds for Serious Applications, Athens, Greece, 2011, pp. 140-143.

[Escriba texto]

[25] S. Brigandi, J. Field, Y. Wang, ´A LEGO Mindstorms NXT Based Multirobot System´, 2010 IEEE/ASME International Conference on Advanced Intelligent Mechatronics Montréal, Canada, 2010, pp. 135-139.

[26] Y. Kim, ´Control Systems Lab using a LEGO Mindstorms NXT Motor System´, 18th Mediterranean Conference on Control & Automation, Marrakech, Morocco, 2010, pp. 173-178.

[27] C.G.L. Cao, E. Danahy, ´Increasing Accessibility to Medical Robotics Education´, in IEEE Conference on Technologies for Practical Robot Applications (TePRA), Medford, MA, USA, 2011, pp. 49-53.

[28] M.M da Silva, J.F. de Magalhães Netto, ´An Educational Robotic Game for Transit Education Based on the Lego MindStorms NXT Platform´, Brazilian Symposium on Computer Games and Digital Entertainment, Manaus, Brazil, 2010, pp.82-87.

[29] E. Cuevas, D. Zaldivar, Pérez-Cisneros, M. 'Low-cost commercial Lego™ platform for mobile robotics´, *International Journal of Electrical Engineering Education*, **47**(2) (2010), 132-150.

[30] Jan Jantzen, Tuning Of Fuzzy PID Controllers, Technical University of Denmark Department of Automation , (1998) Volume: 871, Issue: 98, Pages: 1-22.

[Escriba texto]